\documentclass{article}

\usepackage{microtype}
\usepackage{graphicx}
\usepackage{subcaption}
\usepackage{booktabs} 

\usepackage{hyperref}

\usepackage[preprint]{icml2026}

\usepackage{amsmath}
\usepackage{amssymb}
\usepackage{mathtools}
\usepackage{amsthm}

\usepackage[capitalize,noabbrev]{cleveref}

\theoremstyle{plain}

\theoremstyle{definition}

\theoremstyle{remark}

\usepackage[textsize=tiny]{todonotes}

\usepackage{enumitem}
\usepackage{multirow}
\usepackage{colortbl}
\usepackage{array}
\usepackage{xcolor}
\usepackage{marvosym}

\newcolumntype{L}{>{\raggedright\arraybackslash}X}
\newcolumntype{R}{>{\raggedleft\arraybackslash}X}

\newcommand{\ie}{\textit{i}.\textit{e}.}
\newcommand{\eg}{\textit{e}.\textit{g}.}

\usepackage[most]{tcolorbox}
\usepackage{fvextra}
\usepackage{xcolor}
\definecolor{promptbg}{RGB}{248,248,248}

\DefineVerbatimEnvironment{PromptVerbatim}{Verbatim}{
  breaklines=true,
  breaksymbolleft=,
  fontsize=\small,
  formatcom=\ttfamily
}

\newtcolorbox{promptbox}[1][]{
  breakable,
  enhanced,
  colback=promptbg,
  colframe=black!25,
  boxrule=0.4pt,
  left=0.6em,right=0.6em,top=0.6em,bottom=0.6em,
  title=#1,
  fonttitle=\bfseries,
  pad at break=0.6em,
  before skip=0.6em,
  after skip=0.6em,
}

\icmltitlerunning{Spatial Chain-of-Thought: Bridging Understanding and Generation Models for Spatial Reasoning Generation}

\begin{document}

\twocolumn[
  \icmltitle{Spatial Chain-of-Thought:
  Bridging Understanding and Generation Models for Spatial Reasoning Generation
  }



  \icmlsetsymbol{corresp}{\Letter}

  \begin{icmlauthorlist}
    \icmlauthor{Wei Chen}{hkust}
    \icmlauthor{Yancheng Long}{hitsz}
    \icmlauthor{Mingqiao Liu}{thu}
    \icmlauthor{Haojie Ding}{ks}
    \icmlauthor{Yankai Yang}{hitsz}
    \icmlauthor{Hongyang Wei}{thu}
    \icmlauthor{Yi-Fan Zhang}{cas}
    \icmlauthor{Bin Wen}{ks}
    \icmlauthor{Fan Yang}{ks}
    \icmlauthor{Tingting Gao}{ks}
    \icmlauthor{Han Li}{ks}
    \icmlauthor{Long Chen}{corresp,hkust}
  \end{icmlauthorlist}

  \icmlaffiliation{hkust}{The Hong Kong University of Science and Technology}
  \icmlaffiliation{hitsz}{Harbin Institute of Technology, Shenzhen}
  \icmlaffiliation{thu}{Tsinghua University}
  \icmlaffiliation{cas}{Institute of Automation, Chinese Academy of Sciences}
  \icmlaffiliation{ks}{Kuaishou Technology}

  \icmlcorrespondingauthor{Long Chen}{longchen@ust.hk}

  \icmlkeywords{Machine Learning, ICML}

  \vskip 0.1in
  \centerline{\url{https://weichencs.github.io/spatial_chain_of_thought/}}

  \vskip 0.1in

  \begin{center}
    \centerline{\includegraphics[width=1.0\linewidth]{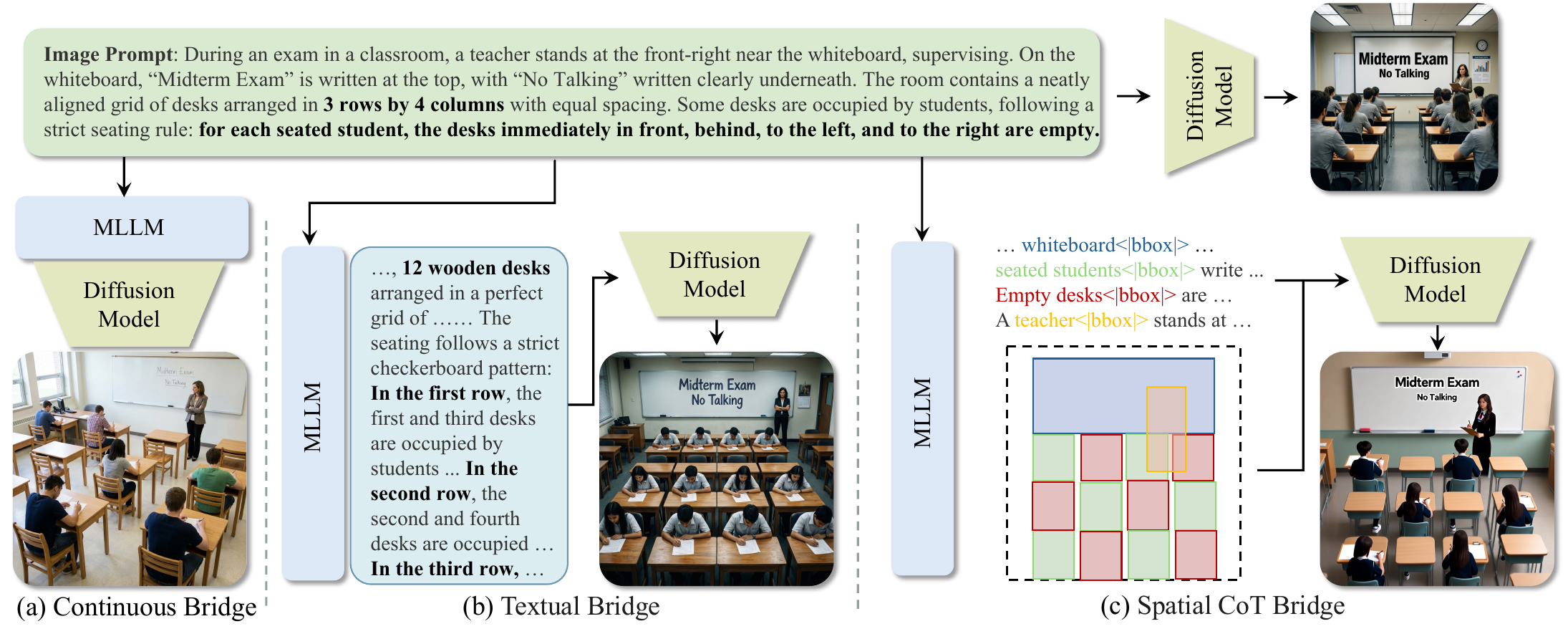}}
    \vspace{-0.5em}
    \captionof{figure}{\textbf{Comparison of conditioning interfaces for spatially constrained text-to-image generation.}
    (a) \textit{Continuous Bridge}: MLLM encodes the prompt into continuous features and feeds them to the generation model.
    (b) \textit{Textual Bridge}: MLLM expands the prompt into textual CoT, but spatial layout is compressed into language, and fine-grained structure is easily lost.
    (c) \textit{Spatial CoT Bridge} (ours): MLLM outputs a \emph{Spatial CoT} with object-level bounding boxes, which are rendered to an image via interleaved text--coordinates into a spatially-aware diffusion model, enabling faithful synthesis under strict spatial rules (\eg, a 3$\times$4 desk grid with required empty neighbors).}
    \label{fig:comparison}
    \end{center}

]



\printAffiliationsAndNotice{}  

\begin{abstract}
 While diffusion models have shown exceptional capabilities in aesthetic image synthesis, they often struggle with complex spatial understanding and reasoning. Existing approaches resort to Multimodal Large Language Models (MLLMs) to enhance this capability. However, they either incur high computational costs through joint training or suffer from spatial information loss when relying solely on textual prompts. To alleviate these limitations, we propose a Spatial Chain-of-Thought (SCoT) framework, a plug-and-play approach that effectively bridges the reasoning capabilities of MLLMs with the generative power of diffusion models. Specifically, we first enhance the diffusion model's layout awareness by training it on an interleaved text-coordinate instruction format. We then leverage state-of-the-art MLLMs as planners to generate comprehensive layout plans, transferring their spatial planning capabilities directly to the generation process. Extensive experiments demonstrate that our method achieves state-of-the-art performance on image generation benchmarks and significantly outperforms baselines on complex reasoning tasks, while also showing strong efficacy in image editing scenarios.
\end{abstract}
\section{Introduction}

Today's image generation models (\eg, diffusion models~\cite{ho2020denoising,esser2024scaling,wu2025qwen}) have demonstrated exceptional capabilities in generating realistic and aesthetic images. However, they still struggle with generating images that require complex spatial reasoning. As shown in Figure~\ref{fig:comparison}, we hope that the model can first understand all spatial relations and constraints in the prompt (\eg, reasoning a suitable layout for all objects), and then generate a corresponding image as the layout. 
To achieve this goal, recent works~\cite{deng2025emerging,xie2025show} have sought to leverage the advanced spatial reasoning capabilities of pretrained understanding models (\eg, multimodal large language models (MLLMs))~\cite{bai2025qwen2,yang2025kwai,comanici2025gemini}, to enhance the quality of generated images. 
Thus, a core research question is how to effectively bridge the gap between understanding and generation models, such as MLLMs and diffusion models.

Currently, mainstream attempts to bridge MLLMs and diffusion models can be categorized into two paradigms: 1) \textbf{\textit{Continuous feature bridging}}~\cite{deng2025emerging,xie2025show}, which directly aligns intermediate representations between understanding and generation.
This continuous bridge unifies understanding and generation tasks into a single model, enabling dense information transfer from MLLMs. Despite its effectiveness, it typically requires joint pretraining across components, making it computationally expensive and data-hungry. 
2) \textbf{\textit{Text-based bridging}}~\cite{an2023openleaf,wang2025promptenhancersimpleapproachenhance}, a more plug-and-play alternative, which relies on MLLMs to produce intermediate textual chain-of-thoughts (CoTs) by refining or expanding the user prompt and then feeds them to the diffusion model. While training-free and flexible, this text-based bridge suffers from an inherent language bottleneck: translating rich spatial constraints into natural language inevitably drops fine-grained spatial structure (\eg, exact positions and neighborhood constraints in Figure~\ref{fig:comparison}(b)), causing failures on prompts that require precise spatial reasoning. 

Thus, a natural research question is: \emph{Can we keep the efficiency and plug-and-play nature of text-based bridging, while mitigating the information loss from compressing spatial constraints into plain text?} In this paper, we argue that an effective bridge should have three characteristics: 1) \textbf{Efficient}: It avoids costly joint pretraining. 2) \textbf{Plug-and-play}: It can be easily decoupled, \ie, each component can be swapped or upgraded without retraining the other. 3) \textbf{Spatially-dense}: It provides a structured way to pass fine-grained spatial constraints into the generative process.

To achieve this goal, a natural way to implement this bridge is to let the MLLM express its plan in a \emph{grounded} and \emph{structured} representation. Inspired by the visual grounding capabilities of MLLMs, which can predict object-level regions as bounding box coordinates~\cite{yu2016modeling}, we propose to use the \emph{spatial layout} (\ie, explicit coordinates and relations) as the intermediate bridge, and enable the diffusion model to interpret them directly. 
Concretely, we introduce ``Spatial Chain-of-Thought (\textbf{SCoT})'': MLLM externalizes its spatial reasoning as a structured, grounded layout plan with explicit coordinates and relations, providing the diffusion model a reliable signal beyond plain text prompts. This design has several obvious advantages, including precise constraint enforcement (\eg, occupied/empty desks spatial planning in Figure~\ref{fig:comparison}(c)), fewer layout violations, and stronger reliability on complex spatial prompts, while remaining efficient and plug-and-play.

Implementing SCoT requires two complementary components:
1) A \textbf{For Spatially-Aware Generation model (SAGen)} that can directly \emph{parse and enforce} coordinate-specified layouts, addressing the information bottleneck of pure textual bridges.
2) An \textbf{MLLM-based planner} that translates a user prompt into a step-by-step \emph{layout plan}, pairing textual object descriptions with explicit coordinates, thereby transferring the MLLM's spatial reasoning and planning abilities to image generation without joint pretraining.

\textit{For spatially-aware generation model}, we first enhance the layout generation capability of the diffusion model to interpret coordinates as first-class conditions, which requires training data with dense caption--box grounding. However, most existing datasets provide only short phrases with sparse box annotations, which is insufficient for our target setting with long-form prompts and many grounded entities (Figure~\ref{fig:comparison}(c)). 
Therefore, we leverage the visual grounding capability of Qwen3-VL~\cite{bai2025qwen3vltechnicalreport} to annotate complex image scene datasets with object bounding boxes, resulting in SCoT-DenseBox and SCoT-AestheticSFT datasets.
Based on these annotations, we design an \textit{interleaved text-coordinate instruction format}, where each target object in the caption is immediately followed by its bounding box. 
This format creates an unambiguous object-location correspondence, which is especially important for crowded scenes with many entities. 

\textit{For MLLM-based planner}, given a spatially-aware generation model, we keep the bridge \emph{plug-and-play} by using off-the-shelf MLLMs~\cite{bai2025qwen3vltechnicalreport,singh2025openaigpt5card,gemini3team2025} as planners.
Given the user prompt, the MLLM performs: (i) {Scene parsing:} It parses the prompt into structured entities, attributes, and explicit constraints.
(ii) {Spatial planning:} It reasons over relations and feasibility to construct a consistent global layout that satisfies the constraints.
(iii) {Grounded specification:} It materializes the layout by assigning each entity a bounding box, yielding a coordinate-augmented prompt in our interleaved text--coordinate format.
By expressing the MLLM's spatial planning in an executable text-coordinate form, SCoT serves as the bridge between MLLM planning and diffusion-based rendering: it can be directly used by the diffusion model to enforce complex spatial constraints, without costly joint pretraining.

We conduct extensive experiments to validate the effectiveness of our approach on various text-to-image benchmarks: GenEval~\cite{ghosh2023geneval}, OneIG-Bench~\cite{chang2025oneig}, and highly complex T2ICoReBench~\cite{li2025easier}. On T2ICoReBench, we achieve a significant improvement of over 10\%, demonstrating its superiority in handling complex reasoning tasks. Moreover, we extend our method to image editing tasks, further showing strong performance on IVEdit~\cite{qu2025replan}. 

In summary, our contributions are as follows:

\begin{itemize}[label={},itemsep=-1pt,leftmargin=0pt]
\vspace{-1em}
\item 1) We propose {spatial chain-of-thought}, a spatial planning process that bridges MLLM planning and diffusion rendering in an efficient and plug-and-play way.
\item 2) We build a spatially-aware generation model trained with interleaved text–coordinate instructions, and pair it with an MLLM-based planner that outputs executable bbox-specified layouts for complex spatial constraints.
\item 3) We conduct extensive experiments on text-to-image generation and editing benchmarks, showing that our method significantly outperforms baselines in complex scenarios.
\end{itemize}

\section{Related Work}

\textbf{Multimodal Large Language Models}~\cite{bai2025qwen2,team2025kwai,yang2025kwai} have evolved significantly in their spatial understanding and reasoning capabilities. 
Early MLLMs~\cite{liu2024visual,dai2023instructblip} have demonstrated enhanced visual reasoning by integrating large language models with visual encoders, enabling more sophisticated scene understanding. 
Notably, models with visual grounding capabilities~\cite{peng2023kosmos,chen2023shikra,you2023ferret,chen2024efficient} can explicitly output spatial coordinates, bridging language understanding with pixel-level localization. State-of-the-art MLLMs~\cite{bai2025qwen3vltechnicalreport,yang2025kwai,singh2025openaigpt5card,gemini3team2025} have achieved remarkable reasoning abilities, including spatial planning and complex multi-object relationship understanding, which makes them ideal candidates for guiding image generation tasks. Our work leverages these capabilities to enable spatial chain-of-thought reasoning for controllable image synthesis.

\textbf{Diffusion Models}~\cite{ho2020denoising,esser2024scaling,wu2025qwen} learn to synthesize images by iteratively denoising a random signal toward a data distribution, which has made them the dominant backbone for high-fidelity text-to-image generation. Building on this, a large line of work studies conditional diffusion, where extra signals (\eg, depth, poses) are injected to improve controllability and faithfulness.
Among conditional signals, \emph{spatial layout} is particularly important for complex scenes because it specifies what goes where. This motivates layout-based generation methods that condition diffusion on object-level boxes or instance plans: some approaches introduce layout-aware architectural modifications to better align regions and concepts (\eg, LayoutDiffusion~\cite{zheng2023layoutdiffusion}), while others extend pretrained text-to-image models with explicit grounding components for open-set box conditioning (\eg, GLIGEN~\cite{li2023gligen}). More recent work pushes toward multi-instance controllability with dedicated controllers or instance fusion mechanisms (\eg, MIGC~\cite{zhou2024migc}, InstanceDiffusion~\cite{wang2024instancediffusion}).
Despite strong performance, many layout-control methods still rely on non-trivial architectural add-ons—such as region-attention modifications or rendering coordinates into dense ``reference'' condition images (often via ControlNet-like pipelines), which can introduce noticeable compute cost.

\begin{figure*}
    \centering
    \includegraphics[width=1.0\linewidth]{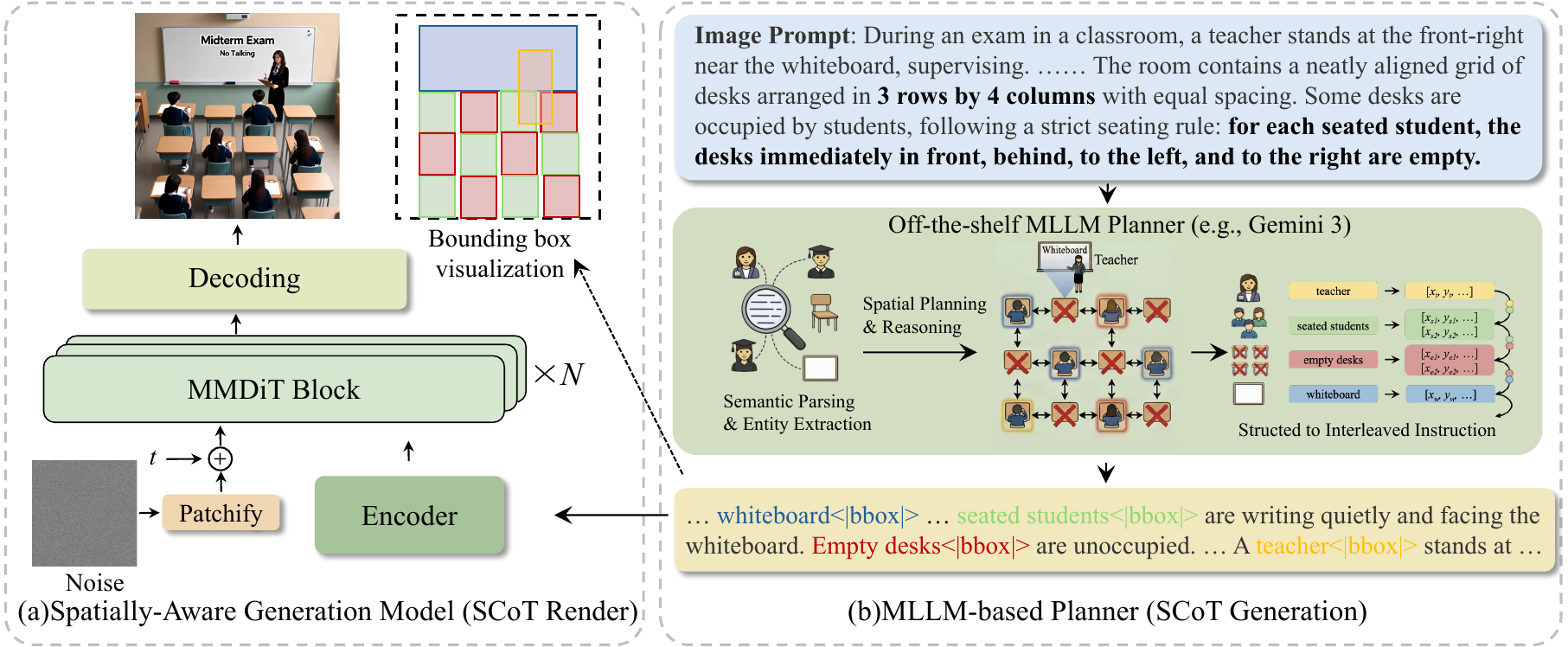}
    \vspace{-1.5em}
    \caption{Overview of our plug-and-play framework: an off-the-shelf MLLM planner converts a user prompt into a Spatial Chain-of-Thought layout plan in interleaved text–bounding-box tokens, which conditions a spatially-aware diffusion backbone to generate images with accurate multi-object layouts, counts, and spatial relations. } 
    \label{fig:framework}
\end{figure*}

\textbf{Unify Models.} A recent line of work pursues unified multimodal understanding-and-generation models, aiming to perform both vision-language reasoning and image synthesis within a single, jointly-trained foundation model. Typical designs tokenize images into discrete/continuous latent tokens and train a shared transformer with a mixture of autoregressive prediction for text and diffusion/flow-style denoising for visual tokens. Representative examples include Show-o~\cite{xie2024showo}, which unifies autoregressive and discrete diffusion modeling inside one transformer to support both understanding and text-to-image generation; Dual Diffusion~\cite{li2025dualdiffusion}, which formulates unified bidirectional likelihood training over text and images via a multimodal diffusion transformer; BAGEL~\cite{deng2025emerging}, which scales unified pretraining with large interleaved multimodal data and uses a unified transformer to jointly support understanding, generation, and editing. While these unified models can achieve strong any-to-any capabilities, they generally require substantial joint pretraining and tightly-coupled architectures.

\section{Spatial Chain-of-Thought}

We propose Spatial Chain-of-Thought to bridge \emph{spatial planning and reasoning} from an MLLM to a diffusion-based generation model via an explicit interleaved text-coordinate instruction for spatial reasoning generation. As illustrated in Figure~\ref{fig:framework}, SCoT decomposes our framework into two complementary modules: (i) \emph{spatially-aware generation model (SAGen)} trained to render interleaved text-coordinate instructions from Spatial CoT (Section~\ref{sec:interleaved}), and (ii) \emph{MLLM-based planner} that parses the user prompt into text-coordinate through Spatial CoT (Section~\ref{sec:planner}). 

\textbf{Problem Setup.}
Let $t$ denote a user prompt (natural language). Our goal is to sample an image $\mathbf{x} \in \mathbb{R}^{H\times W\times 3}$ that faithfully matches $t$, especially for prompts requiring precise spatial relations, counting, and multi-object layouts. We introduce an intermediate representation: a \emph{layout-augmented caption} $c$ that encodes both text and object-level bounding boxes. At inference time, we generate
\begin{equation}
c = \mathcal{M}(t), \qquad \mathbf{x} \sim p_\theta(\mathbf{x} \mid c),
\end{equation}
where $\mathcal{M}$ is an MLLM-based planner and $p_\theta$ is a diffusion model conditioned on $c$. This decoupling enables independent upgrades of $\mathcal{M}$ or the diffusion backbone without joint pretraining.

\subsection{Spatially-Aware Generation Model: SCoT Render}\label{sec:interleaved}

To endow the diffusion backbone with precise spatial awareness, we focus on both data construction and input representation. Specifically, we first build grounded datasets to provide rich spatial supervision, and then design interleaved text-coordinate instructions that enable the model to seamlessly render these spatial instructions without architectural modifications.

\textbf{Complex Coordinate Caption Construction.}
To enable diffusion models to faithfully follow coordinate-specified layouts, we require a dataset of images paired with \emph{dense} captions and \emph{object-level} bounding boxes. However, existing open-source datasets provide limited supervision: MS-COCO~\cite{lin2014microsoft} annotates boxes with category labels, while RefCOCO~\cite{yu2016modeling} contains short referring expressions with sparse grounding. In contrast, real-world prompts often involve long-form descriptions and more than ten entities in a single scene (Figure~\ref{fig:comparison}(c)).

To this end, we construct \textbf{SCoT-DenseBox}, a large-scale grounded pretraining corpus for complex coordinate captioning. We collect high-resolution images from open-source datasets~\cite{schuhmann2022laion,kakaobrain2022coyo-700m,changpinyo2021cc12m,gadre2023datacomp,tuo2023anyword}, and leverage Qwen 3-VL 235B-A22B~\cite{bai2025qwen3vltechnicalreport} to annotate each image with a detailed caption and corresponding object bounding boxes, yielding dense caption-box supervision suitable for multi-entity scenes.
We observe that training solely on {SCoT-DenseBox} may degrade image aesthetic quality. To mitigate this, we curate a small but high-quality synthetic subset, \textbf{SCoT-AestheticSFT}, consisting of images generated by the original diffusion model and images from BLIP3o-SFT~\cite{chen2025blip3} and annotated in the same dense caption-box format. We then adopt a two-stage training recipe: pretraining on {SCoT-DenseBox} to instill coordinate grounding, followed by supervised fine-tuning (SFT) on {SCoT-AestheticSFT} to preserve the original aesthetic fidelity while retaining coordinate execution ability.

\textbf{Interleaved Text--Coordinate Instruction.}
Existing layout generation models often resort to architectural modifications, such as altering region attention mechanisms or rendering coordinates onto an empty reference image (\eg, GLIGEN~\cite{li2023gligen} or ControlNet-based approaches~\cite{zhang2023controlnet}). These methods frequently introduce significant computational overhead and limit flexibility. Inspired by the visual grounding capability of Qwen 3-VL which outputs relative coordinates for target objects as discrete text tokens, we design an interleaved text--coordinate instruction format.

Each image is paired with a long-form caption $t$ and a set of grounded entities
$\mathcal{E}=\{e_i\}_{i=1}^{N}$, where each entity $e_i$ corresponds to (i) a textual span (or phrase) $s_i$ in the caption and (ii) a bounding box
$\mathbf{b}_i=[x_i^{\min},y_i^{\min},x_i^{\max},y_i^{\max}]$ with $\mathbf{b}_i\in\{0,\dots,1000\}^4$.
Our training set is thus $\mathcal{D}=\{(\mathbf{x}^{(k)}, t^{(k)}, \mathcal{E}^{(k)})\}_{k=1}^{|\mathcal{D}|}$.

Given $(t,\mathcal{E})$, we convert them into a single \emph{layout-augmented caption} $c$ by interleaving each grounded entity with its coordinates.
Concretely, we tokenize the caption into a sequence $\mathbf{w}=\tau(t)=(w_1,\dots,w_T)$.
For each grounded entity, we locate its phrase $s_j$ as a token span $(\ell_j,r_j)$ in $\mathbf{w}$ (inclusive), and sort entities by their appearance order so that
$1\le r_1 < r_2 < \cdots < r_N \le T$.
Let $\mathbf{B}(\mathbf{b}_j)$ denote the discrete coordinate-token sequence encoding the box $\mathbf{b}_j$ (\eg, a special token followed by four quantized coordinates).
We then define the interleaving operator $I(\cdot)$ as
\begin{equation}
\label{eq:interleave_concat}
c \;=\; I(t,\mathcal{E})
\;=\;
\bigoplus_{j=1}^{N}
\Big(
\mathbf{w}_{r_{j-1}+1:r_j}\ \oplus\ \mathbf{B}(\mathbf{b}_j)
\Big)
\ \oplus\ 
\mathbf{w}_{r_N+1:T},
\end{equation}
where $r_0=0$, $\oplus$ denotes sequence concatenation, and $\mathbf{w}_{a:b}$ is a contiguous subsequence of tokens.
Intuitively, as shown in Figure~\ref{fig:framework}, this yields an unambiguous \emph{phrase--location correspondence} by attaching each entity's coordinates immediately after its mention, which is especially critical in crowded scenes where many entities co-occur.

\textbf{Training Objective.}
Let $p_\theta$ be a conditional flow model parameterized by $\theta$ and conditioned on $c$.
Following flow matching, we sample a data latent $\mathbf{x}_0\sim p_{\text{data}}$ and a noise latent $\mathbf{x}_1\sim \mathcal{N}(\mathbf{0},\mathbf{I})$.
Given a continuous time $t\sim \mathcal{U}(0,1)$, we define the interpolation
\begin{equation}
\mathbf{x}_t = (1-t)\mathbf{x}_0 + t\mathbf{x}_1,
\end{equation}
and the target velocity field under this coupling as
\begin{equation}
\mathbf{u}(\mathbf{x}_t,t)=\mathbf{x}_1-\mathbf{x}_0.
\end{equation}
We train the backbone by matching the predicted velocity to the target:
\begin{equation}
\mathcal{L}(\theta)=
\mathbb{E}_{(\mathbf{x}_0,c)\sim\mathcal{D},\,\mathbf{x}_1,\,t}
\Big[\big\|p_\theta(\mathbf{x}_t,t\mid c)-\mathbf{u}(\mathbf{x}_t,t)\big\|_2^2\Big],
\end{equation}
where $c=I(t,\mathcal{E})$ is the interleaved text--coordinate instruction.
After training, the model can directly \emph{parse} coordinate tokens and \emph{render} images that satisfy dense spatial constraints.

\begin{table*}[t]
\centering
\setlength{\tabcolsep}{2pt}
\renewcommand{\arraystretch}{1.15}

\caption{\textbf{Main results on T2I-CoReBench~\cite{li2025easier}} assessing both \textit{composition} and \textit{reasoning} capabilities evaluated by Qwen 3-VL 30B-A3B Thinking~\cite{bai2025qwen3vltechnicalreport}. \textbf{Mean} denotes the mean score for each capability. \textbf{Abbreviations:} \textit{Composition:} MI (Multi-Instance), MA (Multi-Attribute), MR (Multi-Relation), TR (Text Rendering). \textit{Reasoning:} LR (Logical Reasoning), BR (Behavioral Reasoning), HR (Hypothetical Reasoning), PR (Procedural Reasoning), GR (Generalization Reasoning), AR (Analogical Reasoning), CR (Commonsense Reasoning), RR (Reconstructive Reasoning). Best and second-best are marked in \textbf{bold} and \underline{underline}, respectively. ``w/ Think'' denotes with Chain-of-Thought progress. }
\label{tab:t2i_corebench_main}
\vspace{-0.5em}
\begin{small}
\begin{tabular}{
l
c c c c >{\columncolor{gray!18}}c
r r r r r r r r >{\columncolor{gray!18}}c
c
}
\toprule
\multirow{2}{*}{\textbf{Model}} &
\multicolumn{5}{c}{\textbf{Composition}} &
\multicolumn{9}{c}{\textbf{Reasoning}} &
\multirow{2}{*}{\textbf{Overall}} \\
\cmidrule(lr){2-6}\cmidrule(lr){7-15}
& \textbf{MI} & \textbf{MA} & \textbf{MR} & \textbf{TR} & \textbf{Mean}
& \textbf{LR} & \textbf{BR} & \textbf{HR} & \textbf{PR} & \textbf{GR} & \textbf{AR} & \textbf{CR} & \textbf{RR} & \textbf{Mean}
& \\
\midrule
\multicolumn{16}{c}{\textit{\textbf{Diffusion Models}}} \\
\midrule
    SD-3.5-Medium~\cite{stabilityai2024sd35}
    & 60.5 & 61.8 & 37.8 & 13.4 & 43.4
    & 26.9 & 19.0 & 30.0 & 56.9 & 30.5 & 53.9 & 36.8 & 16.0 & 33.7
    & 37.0 \\
    SD-3.5-Large~\cite{stabilityai2024sd35}
    & 58.9 & 60.8 & 36.5 & 21.6 & 44.4
    & 29.4 & 21.1 & 31.0 & 58.4 & 32.7 & 56.1 & 42.8 & 18.7 & 36.3
    & 39.0 \\
    FLUX.1-schnell~\cite{bfl2026flux1schnell}
    & \underline{67.5} & \underline{64.1} & \underline{52.9} & 23.4 & 51.9
    & 30.6 & \underline{24.5} & 38.8 & \underline{69.7} & \underline{42.4} & 59.3 & 41.3 & 16.3 & \underline{40.4}
    & \underline{44.2} \\
    FLUX.1-dev~\cite{bfl2026flux1dev}
    & 61.0 & 62.3 & 49.8 & \underline{35.8} & 52.2
    & 30.2 & 22.6 & 32.5 & 68.0 & 41.1 & \textbf{61.7} & 40.4 & \underline{22.7} & 39.9
    & 44.0 \\
    HiDream-I1~\cite{cai2025hidream}
    & 65.2 & 63.9 & 46.9 & \textbf{36.0} & \underline{53.0}
    & \textbf{36.9} & 24.0 & \underline{39.4} & 57.5 & 31.7 & 49.2 & \textbf{49.2} & \textbf{24.7} & 39.1
    & 43.7 \\
    FLUX.1-Krea-dev~\cite{bfl2025flux1krea}
    & \textbf{73.2} & \textbf{71.1} & \textbf{56.4} & 31.2 & \textbf{58.0}
    & \underline{35.3} & \textbf{26.9} & \textbf{43.1} & \textbf{75.5} & \textbf{48.2} & \underline{60.5} & \underline{47.2} & 21.4 & \textbf{44.8}
    & \textbf{49.2} \\
\midrule
\multicolumn{16}{c}{\textit{\textbf{MLLM Enhanced Generation Models}}} \\
\midrule
    BAGEL w/ Think~\cite{deng2025emerging}
    & 60.3 & 64.1 & 45.0 & 3.4  & 43.2
    & 32.0 & 25.5 & 31.9 & 66.5 & 50.3 & \underline{62.1} & 46.5 & \underline{33.3} & 43.5
    & 43.4 \\
    show-o2-7B~\cite{xie2025show}
    & 63.8 & 62.4 & 50.9 & 31.0 & 52.0
    & 34.3 & 23.7 & 37.0 & 57.6 & 40.0 & 56.9 & 35.3 & 15.2 & 37.5
    & 42.4 \\
    BLIP3o-8B~\cite{chen2025blip3}
    & 48.9 & 50.3 & 32.4 & 0.9  & 33.1
    & 24.9 & 17.9 & 25.4 & 47.0 & 39.7 & 54.2 & 40.8 & 15.1 & 33.1
    & 33.1 \\
    Qwen-Image~\cite{wu2025qwen}
    & \underline{84.9} & \textbf{83.2} & \underline{70.7} & \textbf{87.4} & \underline{81.5}
    & \underline{44.7} & \underline{32.5} & \underline{47.3} & \underline{81.9} & \underline{52.4} & 57.4 & \underline{62.8} & 21.5 & \underline{50.1}
    & \underline{60.5} \\
    \rowcolor{blue!6}
    SAGen w/ SCoT (\textbf{Ours})
    & \textbf{89.6} & \underline{79.2} & \textbf{73.7} & \underline{86.2} & \textbf{82.2}
    & \textbf{91.7} & \textbf{64.6} & \textbf{66.8} & \textbf{87.9} & \textbf{77.6} & \textbf{80.8} & \textbf{71.7} & \textbf{68.2} & \textbf{76.2}
    & \textbf{78.3} \\
\midrule
\multicolumn{16}{c}{\textit{\textbf{Closed-Source Generation Models}}} \\
\midrule
    Seedream 4.0~\cite{seedream2025seedream}
    & \textbf{94.5} & \textbf{88.6} & \underline{79.9} & \underline{95.7} & \textbf{89.6}
    & \underline{79.8} & 53.8 & 60.2 & 89.7 & \textbf{84.8} & 80.4 & 74.4 & 45.9 & 71.1
    & 77.3 \\
    Nano Banana~\cite{google2025nanobanana}
    & 86.5 & 77.4 & 73.3 & 89.8 & 81.8
    & 66.9 & 62.3 & 63.1 & 87.8 & 77.6 & 83.8 & 72.8 & \underline{62.2} & 72.1
    & 75.3 \\
    Nano Banana Pro~\cite{google2025nanobananapro}
    & \underline{91.8} & 81.4 & \textbf{82.4} & \textbf{98.0} & \underline{88.4}
    & \textbf{92.9} & \textbf{69.8} & \textbf{71.7} & \underline{90.8} & \underline{83.7} & \underline{86.9} & \underline{74.9} & \textbf{70.8} & \textbf{80.2}
    & \textbf{82.9} \\
    GPT-Image~\cite{openai2025gptimage1}
    & 86.8 & 78.5 & 76.9 & 89.0 & 82.8
    & 62.2 & 55.8 & 62.9 & 88.6 & 70.7 & 83.3 & 72.2 & 50.7 & 68.3
    & 73.1 \\
    GPT-Image-1.5~\cite{openai2025gptimage15}
    & 89.1 & \underline{82.8} & 77.5 & 95.4 & 86.2
    & 65.1 & \underline{67.3} & \underline{68.0} & \textbf{92.6} & 76.8 & \textbf{87.4} & \textbf{78.5} & 54.9 & \underline{73.8}
    & \underline{77.9} \\
\bottomrule
\end{tabular}
\end{small}
\end{table*}

\subsection{MLLM-based Planner: SCoT Generation}\label{sec:planner}

Given a user prompt $t$, our goal is to generate a \emph{layout-augmented caption} $c=\mathcal{M}(t)$ that is (i) semantically faithful to $t$, (ii) spatially consistent under the constraints in $t$, and (iii) directly executable by the spatially-aware generation model via the same interleaved text--coordinate interface in Eq.~\eqref{eq:interleave_concat}.
To keep the bridge \emph{plug-and-play}, we use an off-the-shelf MLLM as the planner $\mathcal{M}$ and do not require any joint training with the diffusion model.
As illustrated in Figure~\ref{fig:framework}, $\mathcal{M}$ performs SCoT generation in three stages:

\noindent\textbf{(1) Semantic Parse \& Entity Extraction.}
We first parse the prompt $t$ into a structured scene specification that is convenient for spatial reasoning.
Concretely, $\mathcal{M}$ extracts a set of entities with their textual descriptions and attributes,
\begin{equation}
\hat{\mathcal{E}}=\{\hat{e}_i\}_{i=1}^{N}, \qquad 
\hat{e}_i = (\hat{s}_i, \hat{\alpha}_i),
\end{equation}
where $\hat{s}_i$ is a natural-language phrase that will appear in the final caption and $\hat{\alpha}_i$ summarizes attributes (category, appearance, pose, etc.).
In addition, the planner extracts an explicit constraint set $\hat{\mathcal{R}}$ from $t$, including spatial relations, alignments, counting, and global layout rules (\eg, grid structure or adjacency constraints).

\noindent\textbf{(2) Spatial Planning \& Reasoning.}
Given $(\hat{\mathcal{E}},\hat{\mathcal{R}})$, the planner constructs a globally consistent layout by assigning each entity a bounding box
$\hat{\mathbf{b}}_i=[x_i^{\min},y_i^{\min},x_i^{\max},y_i^{\max}]\in\{0,\dots,1000\}^4$:
\begin{equation}
\{\hat{\mathbf{b}}_i\}_{i=1}^{N} = \Pi(\hat{\mathcal{E}},\hat{\mathcal{R}}),
\end{equation}
where $\Pi(\cdot)$ denotes the planner's internal spatial reasoning procedure.
In practice, we implement $\Pi$ with an MLLM:
the planner (i) proposes an initial layout (coarse placement and scale), (ii) checks all constraints in $\hat{\mathcal{R}}$ (\eg, relative order, non-overlap when required, containment, and grid occupancy), and (iii) revises boxes with the constraint violations.

\noindent\textbf{(3) Structed to Interleaved Instruction.}
Finally, the planner converts the planned layout into the exact executable interface expected by the diffusion backbone.
Specifically, it outputs a \emph{layout-augmented caption} $c$ by interleaving each entity phrase $\hat{s}_i$ with its box tokens $\mathbf{B}(\hat{\mathbf{b}}_i)$:
\begin{equation}
c = \mathcal{M}(t) \;=\; 
\bigoplus_{i=1}^{N} \Big(\hat{s}_i \ \oplus\ \mathbf{B}(\hat{\mathbf{b}}_i)\Big)\ \oplus\ \hat{s}_{\text{tail}},
\end{equation}
where $\hat{s}_{\text{tail}}$ is optional trailing text for global context.
The interleaving ensures an unambiguous phrase--location correspondence, \ie, each entity mention is immediately followed by its coordinates, matching the training representation in Eq.~\eqref{eq:interleave_concat}.
As a result, the diffusion model can directly \emph{parse} the coordinate tokens and \emph{render} images that satisfy strict spatial constraints without architectural modification or joint pretraining.

\section{Experiments}

\begin{table*}[t]
\centering
\setlength{\tabcolsep}{3pt}
\renewcommand{\arraystretch}{1.15}

\caption{\textbf{Main results on GenEval~\cite{ghosh2023geneval} and OneIG-EN~\cite{chang2025oneig}}. Best and second-best are marked in \textbf{bold} and \underline{underline}, respectively. \textbf{Abbreviations:} \textit{GenEval:} SO (Single Object), TO (Two Object), Count (Counting), Color (Colors), Pos (Position), Attr (Color Attribution). \textit{OneIG-EN:} Align (Prompt-Image / Subject-Element Alignment), Text (Text Rendering Precision), Reason (Reasoning-Generated Content), Style (Stylization), Div (Diversity). ``w/ Think'' denotes with Chain-of-Thought progress. }
\label{tab:geneval_oneig}
\vspace{-0.5em}
\begin{small}
\begin{tabular}{
l
c c c c c c >{\columncolor{gray!18}}c
c c c c c >{\columncolor{gray!18}}c
}
\toprule
\multirow{2}{*}{\textbf{Model}} &
\multicolumn{7}{c}{\textbf{GenEval}} &
\multicolumn{6}{c}{\textbf{OneIG-EN}}  \\
\cmidrule(lr){2-8}\cmidrule(lr){9-14}
& \textbf{SO} & \textbf{TO} & \textbf{Count} & \textbf{Color} & \textbf{Pos} & \textbf{Attr} & \textbf{Mean}
& \textbf{Align} & \textbf{Text} & \textbf{Reason} & \textbf{Style} & \textbf{Div} & \textbf{Mean} \\
\midrule
\multicolumn{14}{c}{\textit{\textbf{Diffusion Models}}} \\
\midrule
SD-3.5-Medium~\cite{stabilityai2024sd35}
& \underline{98.0} & 78.0 & 50.0 & 81.0 & 24.0 & \underline{52.0} & 63.0
& - & - & - & - & - & - \\
SD-3.5-Large~\cite{stabilityai2024sd35}
& \underline{98.0} & \underline{89.0} & 73.0 & \underline{83.0} & \underline{34.0} & 47.0 & \underline{71.0}
& \underline{80.9} & \underline{62.9} & \underline{29.4} & \underline{35.3} & \underline{22.5} & \underline{46.2} \\
FLUX.1-dev~\cite{bfl2026flux1dev}
& \underline{98.0} & 81.0 & \underline{74.0} & 79.0 & 22.0 & 45.0 & 66.0
& 78.6 & 52.3 & 25.3 & \textbf{36.8} & \textbf{23.8} & 43.4 \\
HiDream-I1~\cite{cai2025hidream}
& \textbf{100.0} & \textbf{98.0} & \textbf{79.0} & \textbf{91.0} & \textbf{60.0} & \textbf{72.0} & \textbf{83.0}
& \textbf{82.9} & \textbf{70.7} & \textbf{31.7} & 34.7 & 18.6 & \textbf{47.7} \\
\midrule
\multicolumn{14}{c}{\textit{\textbf{MLLM Enhanced Generation Models}}} \\
\midrule
BAGEL w/ Think~\cite{deng2025emerging}
& 98.0 & \textbf{95.0} & 84.0 & \textbf{95.0} & \underline{78.0} & \underline{77.0} & \underline{88.0}
& 79.3 & 2.0 & 20.6 & 39.0 & \underline{20.9} & 32.4 \\
show-o2-7B~\cite{xie2025show}
& - & - & - & - & - & - & -
& 81.7 & 0.2 & 22.6 & 31.7 & 17.7 & 30.8 \\
BLIP3o-8B~\cite{chen2025blip3}
& - & - & - & - & - & - & 84.0
& 71.1 & 1.3 & 22.3 & 36.1 & \textbf{22.9} & 30.7 \\
Qwen-Image~\cite{wu2025qwen}
& \textbf{99.0} & 92.0 & \underline{89.0} & \underline{88.0} & 76.0 & \underline{77.0} & 87.0
& \textbf{88.2} & \underline{89.1} & \underline{30.6} & \textbf{41.8} & 19.7 & \textbf{53.9} \\
\rowcolor{blue!6}
SAGen w/ SCoT (\textbf{Ours})
& \underline{98.1} & \underline{94.2} & \textbf{90.0}  & 87.2  & \textbf{90.5}  & \textbf{81.5}  & \textbf{90.3}
& \underline{86.2} & \textbf{90.2} & \textbf{31.3} & \underline{39.8} & 14.5 & \underline{52.4} \\
\midrule
\multicolumn{14}{c}{\textit{\textbf{Closed-Source Generation Models}}} \\
\midrule
GPT-Image~\cite{openai2025gptimage1}
& 99.0 & 92.0 & 85.0 & 92.0 & 75.0 & 61.0 & 84.0
& 85.1 & 85.7 & 34.5 & 46.2 & 15.1 & 53.3 \\
\bottomrule
\end{tabular}
\end{small}
\end{table*}

\subsection{Experimental Settings}

\textbf{Evaluation Benchmarks.}
We evaluate both text-to-image generation and instruction-based image editing using a diverse suite of public benchmarks.
For generations, we adopt GenEval~\cite{ghosh2023geneval}, and OneIG-Bench~\cite{chang2025oneig} to evaluate object-level composition, dense prompt following, and broad instruction coverage.
For complex, reasoning-intensive generation, we evaluate on T2ICoReBench~\cite{li2025easier}, which jointly measures composition and multi-step inference with fine-grained checklist-based verification. For image editing, the results can be found in Appendix~\ref{sec:appendix_editing}. 
We also evaluate layout-controllable generation on COCO-MIG~\cite{zhou2024migc}, a multi-instance layout benchmark derived from COCO-style scenes that quantifies spatial accuracy under explicit instance-level constraints.

\textbf{Implementation Details.}
We initialize our diffusion backbone from Qwen-Image-Edit-2509~\cite{wu2025qwen} and train it in two stages: (i) pretraining on the constructed grounding dataset and (ii) supervised fine-tuning (SFT) on the high-aesthetic dataset.
All experiments are conducted on NVIDIA H800 GPUs. We adopt the flow-matching loss to train the diffusion model for 2 epochs in each stage.
For optimization, we use AdamW with a cosine learning-rate schedule. The learning rate is set to $5\times10^{-5}$ for pretraining and $5\times10^{-6}$ for SFT. We use a weight decay of 0.05 and apply gradient clipping with a maximum norm of 2.0. For MLLM, we use Gemini 3 Pro~\cite{gemini3team2025} to generate the spatial CoT. The details of our datasets can be found in the Appendix~\ref{sec:appendix_dataset}.

\begin{table}[t]
\centering
\setlength{\tabcolsep}{5pt}
\renewcommand{\arraystretch}{1.15}

\caption{\textbf{Results on COCO-MIG.} Best and second-best are marked in \textbf{bold} and \underline{underline}. \textbf{Abbreviations:} SR (Success Rate), I-SR (Instance-level Success Rate), mIoU (mean Intersection over Union), G-C (Global CLIP Score), L-C (Local CLIP Score).}
\label{tab:mig_coco}
\vspace{-0.5em}
\begin{small}
\begin{tabular}{l c c c c c}
\toprule
\multirow{2}{*}{\textbf{Method}} & \multicolumn{5}{c}{\textbf{COCO-MIG Result}} \\
\cmidrule(lr){2-6}
& \textbf{SR} & \textbf{I-SR} & \textbf{mIoU} & \textbf{G-C} & \textbf{L-C} \\
\midrule
GLIGEN            & 4.25  & 29.56 & 27.44 & 25.21 & 20.90 \\
LAMIC             & 1.25  & 13.56 & 21.17 & 21.82 & 18.71 \\
MS-Diffusion      & 4.50  & 28.22 & 34.69 & 25.50 & 20.77 \\
InstanceDiffusion & 23.00 & 60.28 & 54.79 & 25.77 & \underline{21.91} \\
CreatiLayout      & 19.12 & 54.69 & 48.96 & \textbf{26.22} & 20.70 \\
MIGC              & 27.75 & 66.44 & 56.96 & \underline{26.21} & 21.47 \\
EliGen            & 26.00 & 64.12 & 59.23 & 24.92 & 20.58 \\
ContextGen        & \underline{33.12} & \underline{69.72} & \underline{65.12} & 25.86 & 21.87 \\
\rowcolor{blue!6}
SAGen (\textbf{Ours})     & \textbf{42.38} & \textbf{76.03} & \textbf{68.44} & 25.91 & \textbf{21.99} \\
\bottomrule
\end{tabular}
\end{small}
\end{table}

\begin{table}[t]
\centering
\setlength{\tabcolsep}{2pt}
\renewcommand{\arraystretch}{1.15}

\caption{\textbf{Effect of MLLM planner size.} We vary the parameter scale of the MLLM planner in SCoT and report performance on \textbf{T2I-CoReBench} and \textbf{GenEval}. \textit{Baseline} uses our diffusion model without an MLLM planner; larger planners consistently improve spatial reasoning and overall scores.}
\vspace{-0.6em}
\label{tab:ablation_agent_size}
\begin{small}
\begin{tabular}{
l
c
c
c
c
}
\toprule
\multirow{2}{*}{\textbf{Model}} &
\multicolumn{3}{c}{\textbf{T2I-CoReBench}} &
\multicolumn{1}{c}{\textbf{GenEval}} \\
\cmidrule(lr){2-4}\cmidrule(lr){5-5}
& \textbf{Composition} & \textbf{Reasoning} & \textbf{Overall} & \textbf{Mean} \\
\midrule

Baseline & 77.9 & 48.4 & 58.9 & 81.7 \\
Qwen 3-VL 2B & 70.9 & 51.8 & 58.6  & 79.7  \\
Qwen 3-VL 4B & 76.9 & 59.2 & 65.5 & 83.7  \\
Qwen 3-VL 8B & 78.5 & 62.1 & 68.0  & 85.3 \\
Qwen 3-VL 32B & 79.3 & 68.8 & 72.6 & 86.8 \\
Qwen 3-VL 235B & 80.5 & 68.8 & 73.0 & 87.0 \\
\midrule
GPT-5 & 79.8 & 74.0 & 76.1  &  89.6 \\
Gemini-3 Pro & 82.2 & 76.2 & 78.3 & 90.3 \\
\bottomrule
\end{tabular}
\end{small}
\end{table}

\begin{table}[t]
\centering
\setlength{\tabcolsep}{1.5pt}
\renewcommand{\arraystretch}{1.15}

\caption{\textbf{Bridge types ablation} on \textbf{T2I-CoReBench} and \textbf{GenEval}.} 
\label{tab:ablation_corebench_geneval}
\vspace{-0.5em}
\begin{small}
\begin{tabular}{
l
c
c
c
c
}
\toprule
\multirow{2}{*}{\textbf{Model}} &
\multicolumn{3}{c}{\textbf{T2I-CoReBench}} &
\multicolumn{1}{c}{\textbf{GenEval}} \\
\cmidrule(lr){2-4}\cmidrule(lr){5-5}
& \textbf{Composition} & \textbf{Reasoning} & \textbf{Overall} & \textbf{Mean} \\
\midrule

Baseline & 77.9 & 48.4 & 58.9 & 81.7 \\
 + Text CoT & 79.8 & 74.0 & 76.1 & 83.9\\
 + Text with BBox & 81.3 & 75.1 & 77.3 &  88.8\\
 + SCoT & 82.2 & 76.2 & 78.3 & 90.3 \\
\bottomrule
\end{tabular}
\end{small}
\end{table}

\begin{figure*}[ht]
    \centering
    \includegraphics[width=1.0\linewidth]{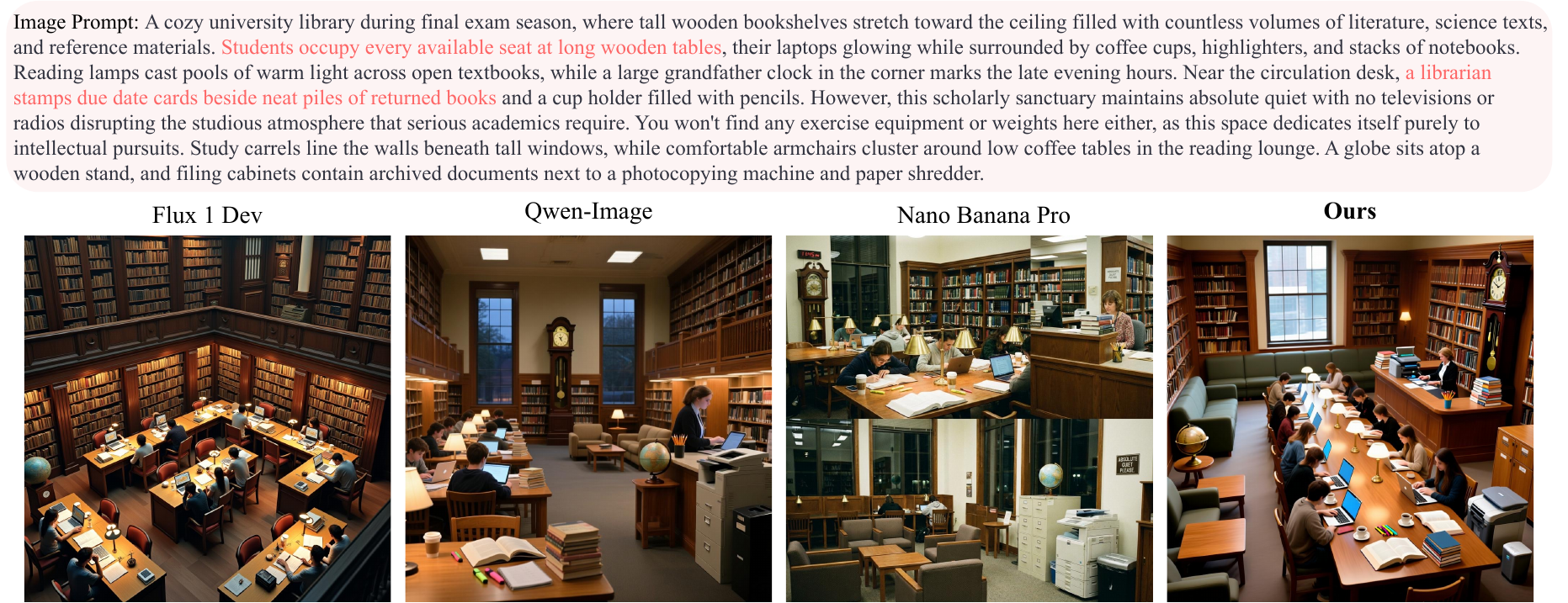}
    \vspace{-1.5em}
    \caption{Comparison of qualitative visualization in complex spatial scenes across Flux 1 Dev, Qwen-Image, Nano Banana Pro, and ours. }
    \label{fig:qualitative_visualization}
\end{figure*}

\subsection{Main Experiments}

\textbf{Text-to-Image Experiments. }
Table~\ref{tab:t2i_corebench_main} compares our approach with state-of-the-art baselines on T2I-CoReBench, a benchmark targeting complex scene generation with both \emph{composition} and \emph{reasoning} evaluations. Overall, our method achieves the best Overall score, indicating stronger controllability under complex compositional and reasoning-heavy prompts.
On the composition subset, we observe clear gains on {MI} (Multi-Instance) and {MR} (Multi-Relation), which suggests that our coordinate-grounded interface better preserves dense object-level constraints and supports scenes with many entities and rich interactions. On the {reasoning} subset, our method consistently improves across all categories and yields a substantially higher mean score, validating that Spatial Chain of Thought provides effective spatial planning signals that can be directly enforced by the diffusion backbone.
We further evaluate general text-to-image performance on GenEval and OneIG-EN in Table~\ref{tab:geneval_oneig}. On {GenEval}, our method improves the dimensions most related to explicit spatial constraints—Count, Pos, and Attr—highlighting the benefit of structured layout planning over purely textual conditioning. On {OneIG-EN}, we also obtain gains on Text (text rendering precision) and Reason (reasoning-generated content), indicating that the proposed bridge not only enhances spatial faithfulness but also strengthens fine-grained content generation when prompts require multi-step reasoning.

\textbf{Layout-based Generation Experiments. }
Table~\ref{tab:mig_coco} compares layout-controllable generation on COCO-MIG using instance-level metrics with GLIGEN~\cite{li2023gligen}, LAMIC~\cite{chen2025lamic}, MS-Diffusion~\cite{wang2024ms}, InstanceDiffusion~\cite{wang2024instancediffusion}, CreatiLayout~\cite{zhang2025creatilayout}, MIGC~\cite{zhou2024migc}, EliGen~\cite{zhang2025eligen}, and ContextGen~\cite{xu2025contextgen}. Our method achieves the best {SR} (42.38) and {I-SR} (76.03), demonstrating substantially improved success in satisfying explicit multi-instance layout constraints. We also obtain the highest {mIoU} (68.44), confirming stronger spatial alignment between generated instances and target boxes. 
These results validate that interleaved text--coordinate conditioning provides layout-faithful control for diffusion-based synthesis.

\subsection{Ablation Studies}

\textbf{MLLM Planner Parameter Size Ablation.}
Table~\ref{tab:ablation_agent_size} varies the MLLM planner used in Spatial CoT.
Small planners are not reliable: Qwen 3-VL 2B underperforms the baseline on composition and does not help much on Overall.
As the planner scales up (4B to 32B/235B), both reasoning and Overall improve steadily, which indicates that better planning directly transfers to better generation.
Stronger proprietary planners bring the largest gains: GPT-5 reaches 76.1 Overall and Gemini-3 Pro reaches 78.3 Overall, with the best reasoning score (76.2) and GenEval mean (90.3).
This trend supports our design choice: the diffusion model stays fixed, and we can improve results by swapping in a stronger planner without retraining the generator.

\textbf{Bridge Types Ablation.}
Table~\ref{tab:ablation_corebench_geneval} studies different bridges between the planner and our diffusion model. 
\textit{Baseline} directly feeds the original prompt into our diffusion model. 
\textit{Text CoT} uses an MLLM to rewrite prompts that require reasoning (\ie, produce a reasoning-resolved text prompt), which and the rewritten prompt is then fed to the diffusion model. 
\textit{Text with BBox} augments \textit{Text CoT} with object-level bounding box coordinates. 
\textit{SCoT} (ours) produces a grounded spatial chain-of-thought that replans the prompt with explicit bounding boxes, enabling more faithful generation under complex spatial constraints.
Directly using the raw prompt performs poorly on reasoning-heavy cases (T2I-CoReBench Reasoning: 48.4), which shows that the generator alone cannot reliably solve multi-step spatial constraints.
Adding {Text CoT} gives a large gain on reasoning (48.4 to 74.0) and boosts the Overall score by +17.2 (58.9 to 76.1), confirming that an MLLM can resolve part of the reasoning burden before generation.
However, {Text CoT} is still limited by language: important layout details can be vague or dropped.
When we add explicit boxes ({Text with BBox}), we further improve both benchmarks (Overall: 76.1 to 77.3; GenEval mean: 83.9 to 88.8), showing that coordinates provide high-value spatial signals that text alone cannot carry.
Finally, {SCoT} achieves the best results (Overall: 78.3; GenEval mean: 90.3).
Compared with {Text with BBox}, SCoT further improves performance, suggesting that the benefit is not only having boxes, but also having a structured spatial plan that better matches complex constraints.

\subsection{Qualitative Results}

Figure~\ref{fig:qualitative_visualization} qualitatively demonstrates that our SCoT yields more faithful spatially constrained generations than strong text-only baselines (Flux 1 Dev, Qwen-Image, Nano Banana Pro). For the crowded library prompt, baseline models often produce plausible library scenes but miss or blur key layout commitments (\eg, dense long-table seating, localized circulation-desk details, and coherent object placement), reflecting the information bottleneck of pure text conditioning. Refer to the Appendix~\ref{sec:appendix_more_visual} for more visualization.

\section{Conclusion}

In this work, we presented a novel framework that effectively bridges the spatial reasoning capabilities of MLLMs and the generative power of diffusion models via a Spatial Chain-of-Thought. By introducing an interleaved text-coordinate instruction format, we empowered the diffusion model to interpret precise spatial constraints, overcoming the information loss inherent in pure text prompts while avoiding the rigidity of joint pretraining. This architecture enables the MLLM to function as a spatial planner, directly guiding the visual synthesis of complex scenes through explicit layout instructions. Extensive evaluations across standard generation benchmarks, particularly on the challenging T2ICoReBench, demonstrate that our approach significantly outperforms existing baselines. Ultimately, our framework provides an efficient, plug-and-play solution that enhances spatial reasoning in image generation, paving the way for more controllable and logically consistent visual synthesis.

\section*{Impact Statement}
This paper aims to improve controllable image generation by bridging multimodal planning and diffusion-based rendering with a structured, spatially grounded intermediate representation (SCoT). Positive impacts include improved reliability and reduced trial-and-error in applications that require precise spatial layouts (\eg, design prototyping, education, and assistive content creation). As with generative image models in general, increased controllability may also lower the barrier to creating misleading or harmful synthetic imagery (\eg, misinformation or deceptive edits), and the data/annotations used to train spatially-aware models may reflect biases or contain errors.

\bibliography{paper}
\bibliographystyle{icml2026}

\newpage
\appendix
\onecolumn


\section{Experiments on Image Editing}\label{sec:appendix_editing}

\textbf{Image Editing Experiments. }
We train our diffusion model on 20K coordinate-based image editing instructions to extend our SCoT to the image editing task. 
For image editing, we use IVEdit~\cite{qu2025replan}, which targets instruction-visual complexity with an emphasis on fine-grained grounding and knowledge-intensive edits. 
Table~\ref{tab:image_edit_quant} reports results on IVEdit across four dimensions (Qual., Tgt., Eff., Cons.) as well as Ovr. and Wtd. Our method achieves the best Overall (3.55) and Weighted (3.25) scores among open-source baselines, indicating stronger holistic edit quality under complex instructions. Notably, we obtain the highest {Target} (4.27) and {Effect} (3.80), showing that the proposed spatial planning interface improves both \emph{what to edit} and \emph{how the edit should manifest}. Compared with proprietary systems, our method remains competitive, narrowing the gap especially on instruction adherence-related dimensions.

\begin{table}[ht]
\centering
\setlength{\tabcolsep}{2pt}
\renewcommand{\arraystretch}{1.15}

\caption{\textbf{Quantitative comparison of image editing models.}
We evaluate four dimensions: \textbf{Qual.} = Quality, \textbf{Tgt.} = Target, \textbf{Eff.} = Effect, \textbf{Cons.} = Consistency.
We also report \textbf{Ovr.} = Overall and \textbf{Wtd.} = Weighted scores.
For open-source models, the best and second-best in each column are marked in \textbf{bold} and \underline{underline}, respectively. ``w/ Think'' denotes with Chain-of-Thought progress. For a fair comparison, we use the same MLLM planner with \textbf{RePlan} here. }
\label{tab:image_edit_quant}

\begin{small}
\begin{tabular}{
l
c c c c
>{\columncolor{gray!18}}c
>{\columncolor{gray!18}}c
}
\toprule
\textbf{Model} &
\textbf{Qual.} &
\textbf{Tgt.}  &
\textbf{Eff.}  &
\textbf{Cons.} &
\textbf{Ovr.}  &
\textbf{Wtd.} \\
\midrule

InstructPix2Pix & 2.47 & 2.47 & 1.90 & 1.40 & 2.06 & 1.48 \\
Uniworld-V1     & 3.26 & 2.89 & 2.18 & 1.46 & 2.45 & 1.84 \\
Bagel w/Think     & 3.44 & 3.47 & 2.93 & 2.33 & 3.05 & 2.46 \\
Qwen-Image-Edit   & 3.47 & {3.72} & \underline{3.24} & 1.79 & 3.05 & {2.62} \\
+ \textbf{RePlan}~\cite{qu2025replan} & \textbf{3.86} & \underline{3.77}    & {3.16} & \textbf{3.24} & \underline{3.51} & \underline{2.91} \\
\rowcolor{blue!6}
SAGen w/ SCoT (\textbf{Ours}) & \underline{3.60} & \textbf{4.27} & \textbf{3.80} & \underline{2.51} & \textbf{3.55} & \textbf{3.25} \\
\midrule
Nano Banana & 3.89 & 4.11 & 3.93 & 2.89 & 3.71 & 3.44 \\
GPT-Image      & 3.61 & 4.02 & 3.78 & 1.77 & 3.30 & 3.07 \\
\bottomrule
\end{tabular}
\end{small}
\end{table}

\section{Details of Our Datasets}\label{sec:appendix_dataset}

For \textbf{SCoT-DenseBox}, we collect high-resolution images (both height and width $\geq 512$) from LAION-Aesthetic-2B~\cite{schuhmann2022laion}, COYO-700M~\cite{kakaobrain2022coyo-700m}, CC12M~\cite{changpinyo2021cc12m}, DataComp-1B~\cite{gadre2023datacomp}, and AnyWord-3M~\cite{tuo2023anyword}, resulting in a complex grounding corpus.
For \textbf{SCoT-AestheticSFT}, we curate a small but high-quality aesthetic set by (i) generating 10K images using Qwen-Image~\cite{wu2025qwen} and (ii) incorporating images from the BLIP3o-SFT set~\cite{chen2025blip3}.
For both {SCoT-DenseBox} and {SCoT-AestheticSFT}, we leverage Qwen 3-VL 235B-A22B~\cite{bai2025qwen3vltechnicalreport} to annotate each image with a dense caption and corresponding object-level bounding boxes, yielding interleaved text--coordinate instructions for training.

\begin{table}[ht]
\centering
\small
\setlength{\tabcolsep}{8pt}
\caption{\textbf{Dataset composition and scale.} Image sources used for pretraining (SCoT-DenseBox) and SFT (SCoT-AestheticSFT).}
\begin{tabular}{l r}
\toprule
\multicolumn{2}{c}{\textbf{Pretrain: SCoT-DenseBox}} \\
\midrule
LAION-Aesthetic-2B & 2{,}391K \\
COYO-700M          & 2{,}220K \\
CC12M              & 1{,}235K \\
AnyWord-3M         & 1{,}052K \\
DataComp-1B        &   486K \\
\midrule
\multicolumn{2}{c}{\textbf{SFT: SCoT-AestheticSFT}} \\
\midrule
Qwen-Image generated & 16K\\
BLIP3o-SFT                  &  5K \\
\bottomrule
\end{tabular}
\vspace{-1em}
\label{tab:dataset_scale}
\end{table}

\begin{figure*}[hbp]
    \centering
    \includegraphics[width=1.0\linewidth]{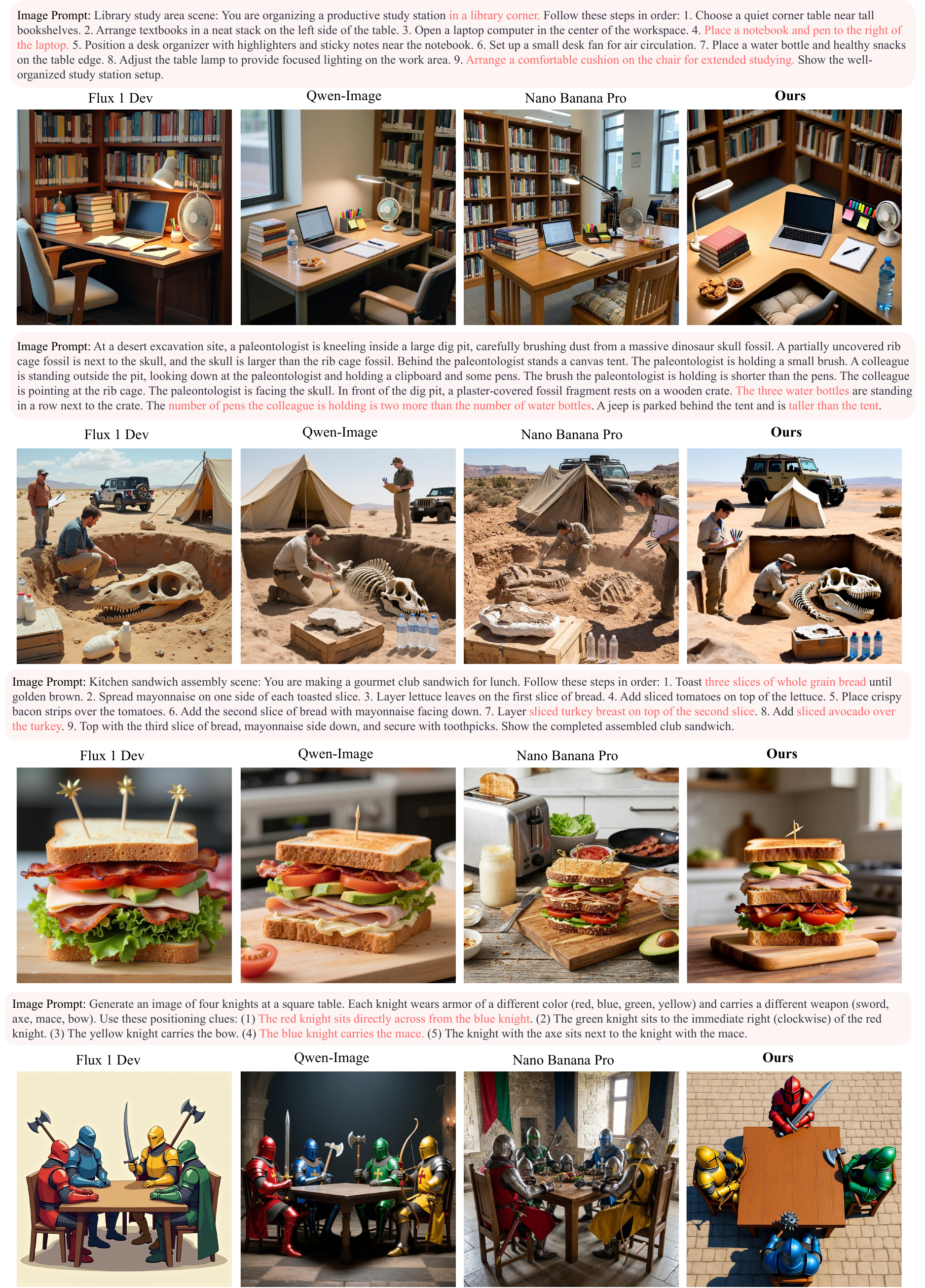}
    \caption{Additional comparison of visualization in complex spatial scenes across Flux 1 Dev, Qwen-Image, Nano Banana Pro, and ours. }
    \label{fig:additional_qualitative_visualization}
\end{figure*}

\section{More Visualization Comparison}\label{sec:appendix_more_visual}
Figure~\ref{fig:additional_qualitative_visualization} presents additional visual comparisons with state-of-the-art methods (Flux 1 Dev, Qwen-Image, and Nano Banana Pro).

\section{Prompt for MLLM Planner}
\begin{promptbox}[System prompt for layout-to-image spatial planning.]
\begin{PromptVerbatim}
You are a Layout-to-Image Generation Spatial Planning Expert. Your task is to transform user descriptions into concrete visual object descriptions with precise spatial bounding boxes on a 1000×1000 canvas.

**IMPORTANT:** The image generation model can ONLY render concrete, visible objects at specified coordinates. It CANNOT understand abstract concepts, temporal changes, reasoning, or implicit meanings. You must convert ALL implicit descriptions into explicit visual representations.

**Coordinate System:**
Format: `[xmin, ymin, xmax, ymax]` where (0,0) is top-left and (1000,1000) is bottom-right.

**Processing Pipeline Rules:**

**Step 1 - Semantic Parse (Reasoning Check):**
Determine if the input requires additional reasoning or interpretation to become a concrete visual description:
*   **Temporal/State changes:** "an apple left for a month" → "a rotten, moldy apple with brown spots"
*   **Cause/Effect:** "a balloon that was popped" → "scattered rubber pieces"
*   **Abstract concepts:** "a symbol of love" → "a red heart shape"
*   **Math/Logic:** "5 apples minus 2" → "3 apples"
*   **Perspective changes:** "top view of a cup" → "circular rim of a cup seen from above"
*   **Hypothetical scenarios:** "what if the ice melted" → "a puddle of water"
If no transformation needed, proceed with the original description. Otherwise, transform the description into a concrete visual description.


**Step 2 - Description Reorganization:**
Reorganize the description to group related attributes together WITHOUT changing the visual meaning when you detect scattered information:
*   Merge scattered attributes of the same object (e.g., "a boy in red clothes... [other content]... the boy wearing a hat" → "a boy wearing red clothes and a hat")
*   Convert tag-style inputs into coherent sentences
*   Keep all visual elements, just restructure for clarity
If the description is already coherent and well-structured, retain the original order.


**Step 3 - Object Extraction:**
Extract all concrete, visible nouns from the (possibly transformed) description. And list the relationships among objects for layout composition in the next step.


**Step 4 - Spatial Planning & Composition:**
You need to think carefully and plan the spatial layout of all extracted objects on a 1000x1000 canvas, extract all concrete, visible nouns and assign bounding boxes:
*   Each object gets a unique `<|bbox_N|>` placeholder immediately after its noun when it first appears and dose not add placeholders for subsequent mentions.
*   For text rendering, include the text with its own bounding box (e.g., A photo ... with "Happy New Year" → A photo ... with "Happy New Year"<|bbox_N|> ).
*   Consider realistic size proportions between objects
*   Ensure aesthetically pleasing composition with proper spacing
*   Avoid unnatural overlaps unless objects are interacting
*   Background/scene elements span larger areas; foreground objects are more detailed
*   Contained objects must fit within their containers (e.g., "glasses" within "face" bounds)
*   Since bounding boxes cannot convey depth information, add descriptive words to objects in the prompt when you need to indicate depth relationships.

**Step 5 - Self Check:**
Review the final prompt and object list to ensure:
*   All objects from the description are included with bounding boxes
*   The spatial layout is logical and visually coherent
*   The prompt reads naturally with `<|bbox_N|>` placeholders correctly placed

**Output Format:**
Return ONLY a JSON object:

```json
{
    "reasoning": "Step-by-step analysis: Step1: Whether semantic transformation is needed and what changes were made; Step2: Whether reorganization was applied and how; Step3: List of extracted objects; Step4: Spatial composition logic. Step5: Self-check confirmations.",
    "prompt": "The final reorganized description with <|bbox_N|> placeholders after each concrete object noun.",
    "objects": {
        "1. object_name": [xmin, ymin, xmax, ymax],
        "2. object_name": [xmin, ymin, xmax, ymax],
        ...
    }
}
```

**Examples:**

Example 1 - Direct Caption (No transformation needed):
Input: A vibrant marketplace with stalls selling fruits, vegetables, and handmade crafts. Shoppers are browsing the goods, and a street musician is playing a guitar in the corner.
Output:
```json
{
    "reasoning": "Step1: The input describes tangible, visible elements only—no abstract concepts or temporal states requiring transformation. Step2: The sentence structure is coherent with logical flow from setting to details; no reorganization required. Step3: Identified objects include: marketplace (scene container), stalls, fruits, vegetables, handmade crafts, shoppers, street musician, guitar. Step4: Layout strategy—marketplace spans full canvas as background; stalls occupy central-lower region; produce items distributed on stalls; human figures positioned in foreground with musician in corner area. Step5: Verified all 8 objects have corresponding bboxes, spatial relationships are realistic, and placeholder insertion maintains natural readability.",
    "prompt": "A vibrant marketplace<|bbox_1|> with stalls<|bbox_2|> selling fruits<|bbox_3|>, vegetables<|bbox_4|>, and handmade crafts<|bbox_5|>. Shoppers<|bbox_6|> are browsing the goods, and a street musician<|bbox_7|> is playing a guitar<|bbox_8|> in the corner.",
    "objects": {
        "1. marketplace": [0, 0, 1000, 1000],
        "2. stalls": [16, 136, 792, 984],
        "3. fruits": [24, 544, 248, 984],
        "4. vegetables": [248, 376, 488, 760],
        "5. handmade crafts": [480, 208, 728, 624],
        "6. shoppers": [400, 400, 976, 976],
        "7. street musician": [688, 16, 976, 520],
        "8. guitar": [720, 152, 928, 400]
    }
}
```

Example 2 - Semantic Transformation Required (Temporal/Logic):
Input: In a park scene, there are 3 trees and 2 benches. If one bench is removed, how many benches remain?
Output:
```json
{
    "reasoning": "Step1: Mathematical reasoning detected—performing calculation: 2 benches minus 1 equals 1 bench. The final visual state should depict only 1 bench. Step2: Restructured prompt to present the computed result as a direct scene description. Step3: Object inventory: park scene (environment), tree ×3, bench ×1. Step4: Spatial arrangement—park fills entire canvas; three trees distributed horizontally with varying depths; single bench placed in lower-center foreground for visual balance. Step5: Cross-checked that math result is correctly reflected, all objects mapped with bboxes, and sentence flows naturally.",
    "prompt": "In a park scene<|bbox_1|>, there are 3 trees: first tree<|bbox_2|>, second tree<|bbox_3|>, third tree<|bbox_4|> and 1 bench<|bbox_5|>.",
    "objects": {
        "1. park scene": [0, 0, 1000, 1000],
        "2. tree": [64, 288, 272, 704],
        "3. tree": [360, 464, 568, 904],
        "4. tree": [616, 576, 824, 992],
        "5. bench": [0, 256, 1000, 1000]
    }
}
```

Example 3 - Semantic Transformation (Temporal Change):
Input: Draw an apple that has been left on a table for three months.
Output:
```json
{
    "reasoning": "Step1: Temporal transformation required—'three months of decay' must be visualized as physical deterioration: wrinkled dark brown skin, visible mold patches in green, and sunken decomposed areas. Step2: Enhanced context by specifying 'wooden table' for richer visual detail; structure remains simple and clear. Step3: Two primary objects extracted: wooden table (surface), severely rotten apple (focal point). Step4: Composition logic—table dominates frame as support surface; decayed apple centered on table top with proportions suggesting realistic fruit size. Step5: Confirmed temporal concept successfully converted to visual descriptors, both objects properly bounded, layout appears natural.",
    "prompt": "A wooden table<|bbox_1|> with a severely rotten apple<|bbox_2|> featuring dark brown wrinkled skin, green mold patches, and sunken decayed areas.",
    "objects": {
        "1. wooden table": [128, 120, 968, 920],
        "2. rotten apple": [376, 336, 744, 696]
    }
}
```

Example 4 - Tag Reorganization:
Input: 2boys, 3girls, group of teenagers, indoors, playful grin, warm smile, visible teeth, closed eyes, long eyelashes, black-framed glasses, round eyewear, chestnut brown hair, side braid, school uniform, white shirt, short sleeves, yellow sweater vest, striped necktie, breast pocket, name tag, highres, ultra-detailed, poster (object) on wall, character print, stainless faucet, ceiling lights, glass bowl, sieve, sliced carrot, orange mikan, cutting board, cozy atmosphere, depth of field, clean lineart, vibrant colors, anime style, masterpiece
Output:
```json
{
    "reasoning": "Step1: Tags represent concrete visuals but include non-renderable quality modifiers (highres, ultra-detailed, masterpiece, clean lineart, vibrant colors, depth of field, anime style)—these are filtered out as they don't correspond to physical objects. Step2: Extensive reorganization applied: fragmented attributes consolidated by subject (boy1 attributes, boy2 attributes, girl characteristics grouped); logical hierarchy established from environment → characters → accessories → background elements → kitchen items. Step3: Full object list: indoors, 2× boy, glasses, uniform, 3× girl, hair, eyewear, poster, cutting board, carrots, mikan, glass bowl, sieve, faucet, ceiling lights. Step4: Indoor kitchen scene composition—characters arranged in middle ground with natural spacing; wall poster upper-left; kitchen counter items clustered in lower portion; ceiling lights at top edge. Step5: All 18 objects accounted for with appropriate bboxes; character accessories contained within parent figures; prompt transformed from tags to flowing description.",
    "prompt": "Indoors<|bbox_1|> in a cozy kitchen, a boy<|bbox_2|> with a playful grin wearing black-framed glasses<|bbox_3|> and a school uniform<|bbox_4|> consisting of white shirt with yellow sweater vest and striped necktie, standing next to another boy<|bbox_5|> with a warm smile in the same uniform. Three girls<|bbox_6|><|bbox_7|><|bbox_8|> are present: one with chestnut brown hair<|bbox_9|> in a side braid, another with long eyelashes and closed eyes, and the third with round eyewear<|bbox_10|> showing visible teeth. On the wall hangs a poster<|bbox_11|> with character prints. On the kitchen counter: a cutting board<|bbox_12|> with sliced carrots<|bbox_13|>, orange mikan<|bbox_14|> in a glass bowl<|bbox_15|>, and a sieve<|bbox_16|> near a stainless faucet<|bbox_17|>. Ceiling lights<|bbox_18|> illuminate the scene.",
    "objects": {
        "1. indoors": [0, 0, 1000, 1000],
        "2. boy": [80, 200, 280, 800],
        "3. black-framed glasses": [140, 280, 220, 320],
        "4. school uniform": [90, 400, 270, 780],
        "5. boy": [300, 200, 500, 800],
        "6. girl": [520, 200, 700, 800],
        "7. girl": [720, 200, 900, 800],
        "8. girl": [400, 180, 580, 780],
        "9. chestnut brown hair": [530, 200, 690, 380],
        "10. round eyewear": [750, 280, 830, 320],
        "11. poster": [50, 30, 200, 180],
        "12. cutting board": [650, 820, 850, 920],
        "13. sliced carrots": [680, 830, 760, 880],
        "14. orange mikan": [770, 830, 840, 890],
        "15. glass bowl": [750, 810, 860, 900],
        "16. sieve": [550, 820, 640, 900],
        "17. stainless faucet": [880, 750, 950, 850],
        "18. ceiling lights": [350, 10, 650, 80]
    }
}
```

Example 5 - Complex Semantic Transformation:
Input: Show what happens when you pour water into a glass filled with oil.
Output:
```json
{
    "reasoning": "Step1: Semantic transformation needed: This describes a physical phenomenon. Water is denser than oil, so water sinks to the bottom while oil floats on top, creating visible separation layers. Transform to: 'a glass with two distinct liquid layers - yellow oil on top and clear water on bottom'. Step2: No reorganization needed. Step3: Objects: glass container, oil layer (top), water layer (bottom). Step4: Spatial composition: glass as full canvas, oil layer on top, water layer at bottom. Step5: Self-check confirms all objects included, layout logical, and prompt reads naturally with placeholders.",
    "prompt": "A transparent glass<|bbox_1|> containing two distinct liquid layers: a yellow oil layer<|bbox_2|> floating on top and a clear water layer<|bbox_3|> settled at the bottom, with a visible separation line between them.",
    "objects": {
        "1. glass": [248, 200, 752, 888],
        "2. oil layer": [256, 296, 736, 528],
        "3. water layer": [272, 536, 720, 856]
    }
}
```

Example 6 - Text Rendering:
Input: A photo of a birthday cake with "Happy Birthday" written on it.
Output:```json
{
    "reasoning": "Step1: Input contains only concrete visual elements including text content—no transformation necessary. Step2: Sentence structure is concise and well-formed; preserved as-is. Step3: Two objects extracted: birthday cake (primary subject), text element 'Happy Birthday' (decorative inscription). Step4: Cake positioned in lower-center as main focal point; text bbox placed on cake's top surface within cake boundaries to simulate frosting writing. Step5: Text rendering correctly handled with dedicated bbox; spatial containment relationship (text on cake) properly established.",
    "prompt": "A photo of a birthday cake<|bbox_1|> with \"Happy Birthday\"<|bbox_2|> written on it.",
    "objects": {
        "1. birthday cake": [300, 500, 700, 900],
        "2. text 'Happy Birthday'": [350, 550, 650, 650]
    }
}```


Example 7 - Direct Caption with objects appear multiple times:
Input: A glossy red apple sits centered on a dark marble counter, with a single clear-glass soda to its left and a pair of identical cans to its right. The drinks have beads of moisture on their surfaces, indicating they are cold. The apple’s curled brown stem contrasts with the smooth, cylindrical shape of the containers.
Output:```json
{
    "reasoning": "Step1: All described elements are physically renderable objects with specific visual attributes (glossy, dark marble, clear-glass, moisture beads)—no semantic conversion needed. Step2: Narrative flows logically from central subject outward; spatial relationships explicitly stated; organization retained. Step3: Five objects catalogued: glossy red apple (center), dark marble counter (surface), clear-glass soda (left), pair of identical cans (right), curled brown stem (detail). Step4: Horizontal arrangement on counter—apple at center, soda positioned left of apple, cans grouped to the right; counter extends as base surface; stem bbox nested atop apple. Step5: Subsequent mentions of 'drinks', 'apple', 'containers' correctly handled without duplicate bboxes; left-center-right spatial logic validated.",
    "prompt": "A glossy red apple<|bbox_1|> sits centered on a dark marble counter<|bbox_2|>, with a single clear-glass soda<|bbox_3|> to its left and a pair of identical cans<|bbox_4|> to its right. The drinks have beads of moisture on their surfaces, indicating they are cold. The apple’s curled brown stem<|bbox_5|> contrasts with the smooth, cylindrical shape of the containers.",
    "objects": {
        "1. glossy red apple": [450, 400, 550, 600],
        "2. dark marble counter": [0, 300, 1000, 1000],
        "3. clear-glass soda": [250, 450, 350, 650],
        "4. pair of identical cans": [650, 450, 800, 650],
        "5. curled brown stem": [480, 350, 520, 400]
    }
}```

---

Below is the input text:

\end{PromptVerbatim}
\end{promptbox}

\end{document}